\documentclass[conference]{IEEEtran}

\makeatletter

\def\ps@IEEEtitlepagestyle{%
  \def\@evenfoot{}%
}

\def\mycopyrightnotice{%
  {\footnotesize 978-1-6654-4231-2/21/\$31.00~\copyright~2021 IEEE\hfill}% <--- Change here
  \gdef\mycopyrightnotice{}
}

\usepackage{blindtext}
\usepackage{eso-pic}
\IEEEoverridecommandlockouts
\usepackage{cite}
\usepackage{amsmath,amssymb,amsfonts}
\usepackage{algorithmic}
\usepackage{graphicx}
\usepackage{textcomp}
\usepackage{xcolor}
\def\BibTeX{{\rm B\kern-.05em{\sc i\kern-.025em b}\kern-.08em
    T\kern-.1667em\lower.7ex\hbox{E}\kern-.125emX}}
    
\usepackage{eso-pic}
\usepackage[hang,small,bf]{caption}
\usepackage[subrefformat=parens]{subcaption}
\usepackage{multirow}
\usepackage{comment}
\usepackage{url}
\AtBeginDocument{
  \abovedisplayskip     =0.4\abovedisplayskip
  \abovedisplayshortskip=0.4\abovedisplayshortskip
  \belowdisplayskip     =0.4\belowdisplayskip
  \belowdisplayshortskip=0.4\belowdisplayshortskip}

\begin{document}
\onecolumn
© 2021 IEEE. Personal use of this material is permitted. Permission from IEEE must be obtained for all other uses, in any current or future media, including reprinting/republishing this material for advertising or promotional purposes, creating new collective works, for resale or redistribution to servers or lists, or reuse of any copyrighted component of this work in other works.
\twocolumn
\newpage
\title{\vspace*{1cm} Constrained Generalized Additive 2 Model With Consideration of High-Order Interactions\\}

\author{\IEEEauthorblockN{1\textsuperscript{st} Akihisa Watanabe}
\IEEEauthorblockA{\textit{dept. of industrial engineering and}\\ 
\textit{economics}\\
\textit{Tokyo Institute of Technology} \\
Tokyo, Japan \\
watanabe.a.az@m.titech.ac.jp}
\and
\IEEEauthorblockN{2\textsuperscript{nd} Michiya Kuramata}
\IEEEauthorblockA{\textit{dept. of industrial engineering and}\\ 
\textit{economics}\\
\textit{Tokyo Institute of Technology} \\
Tokyo, Japan \\
kuramata.m.aa@m.titech.ac.jp}
\and
\IEEEauthorblockN{3\textsuperscript{rd} Kaito Majima}
\IEEEauthorblockA{\textit{dept. of industrial engineering and}\\ 
\textit{economics}\\
\textit{Tokyo Institute of Technology} \\
Tokyo, Japan \\
majima.k.aa@m.titech.ac.jp}
\and
\IEEEauthorblockN{4\textsuperscript{th} Haruka Kiyohara}
\IEEEauthorblockA{\textit{dept. of industrial engineering and}\\ 
\textit{economics}\\
\textit{Tokyo Institute of Technology} \\
Tokyo, Japan \\
kiyohara.h.aa@m.titech.ac.jp}
\and
\IEEEauthorblockN{5\textsuperscript{th} Kondo Kensho}
\IEEEauthorblockA{\textit{dept. of industrial engineering and}\\ 
\textit{economics}\\
\textit{Tokyo Institute of Technology} \\
Tokyo, Japan \\
kondo.k.aq@m.titech.ac.jp}
\and
\IEEEauthorblockN{6\textsuperscript{th} Kazuhide Nakata}
\IEEEauthorblockA{\textit{dept. of industrial engineering and}\\ 
\textit{economics}\\
\textit{Tokyo Institute of Technology} \\
Tokyo, Japan \\
nakata.k.ac@m.titech.ac.jp}
}

\maketitle
%\conf{\textit{  Proc. of the International Conference on Electrical, Computer and Energy Technologies (ICECET) \\ 
%9-10 December 2021, Cape Town-South Africa}}
\begin{abstract}
In recent years, machine learning and AI have been introduced in many industrial fields. In fields such as finance, medicine, and autonomous driving, where the inference results of a model may have serious consequences, high interpretability, as well as prediction accuracy is required. In this study, we propose Constrained Generalized Additive 2 Model plus (CGA\textsuperscript{2}M+), which is based on the Generalized Additive 2 Model (GA\textsuperscript{2}M) and differs from it in two major points. The first is the introduction of monotonicity. Enforcing monotonicity constraints on some features using an analyst's knowledge, is expected to improve not only interpretability but also generalization performance. The second is the introduction of a higher-order term: given that GA\textsuperscript{2}M considers only up to second-order interactions, we introduce a higher-order term, which can capture higher-order interactions and improve prediction performance. In this process, the higher-order term does not reduce interpretability by applying our training method. Numerical experiments showed that the proposed model has high prediction accuracy and interpretability. Furthermore, we confirmed that generalization performance is improved by imposing monotonicity constraints.
\end{abstract}

%\copyrightnotice{XXX-X-XXXX-XXXX-X/XX/\$XX.00 ©20XX IEEE}


\section{Introduction}
In recent years, advances in machine learning and improvements in computer performance have made it possible to make highly accurate predictions for a variety of tasks. However, there is generally a trade-off between prediction accuracy and interpretability, and methods that have high prediction accuracy such as neural networks and gradient boosting decision trees (GBDTs) tend to have low interpretability. However, as stated in the EU’s “General Data Protection Regulation” \cite{EU} and Japan’s “Draft Guidelines for AI Development” \cite{japan}, the interpretability and accountability of AI are becoming more important. In particular, in areas where the results of machine learning inferences can have serious consequences, such as finance, medicine, and autonomous driving, both interpretability and prediction accuracy are even more important. Thus, prediction models are needed that achieve both high prediction accuracy and interpretability.

There are two main directions for interpretability. The first is local interpretation of inference results. In this approach, the inference results of a machine learning models are interpreted individually, such as LIME \cite{LIME} and SHAP \cite{SHAP}. The second is global interpretation of the model, in which the model itself is structured such that it can be interpreted globally. In this study, we build a model for global interpretability from the viewpoint of generality.

The outline of this paper is as follows. In Section II, we present an overview of related works. In Section III, we discuss the problems of the Generalized Additive 2 Model (GA\textsuperscript{2}M), and explain the proposed method, Constrained Generalized Additive 2 Model plus (CGA\textsuperscript{2}M+), which addresses the problems of GA\textsuperscript{2}M. In Section IV, we describe numerical experiments conducted to verify the validity of the improvements and compare the prediction accuracy of CGA\textsuperscript{2}M+ and existing models. In addition, we introduce some examples of applications that take advantage of the interpretability of CGA\textsuperscript{2}M+. Finally, our conclusions and a summary are given in Section V.

\section{Related Work}
In this section, we divide existing machine learning models into the following three categories:
\begin{quote}
 \begin{itemize}
  \item Models that focus on prediction accuracy
  \item Models that focus on interpretability
  \item Balanced models
 \end{itemize}
\end{quote}

First, we introduce support vector machine (SVM) \cite{SVM}, random forests \cite{RF}, the GBDT \cite{GBDT}, and neural networks as models that focus on prediction accuracy. An SVM is a model based on margin maximization. In combination with a kernel method, an SVM can capture high-dimensional nonlinearities and thus generally has high prediction accuracy. A random forest is an ensemble method in which multiple decision trees are trained in parallel using bootstrap sampling, and it generally has higher generalization performance than a normal decision tree. A GBDT is a model for training multiple decision trees sequentially and finally construct an ensemble of them. In particular, LightGBM \cite{LGBM}, which has been designed to reduce time complexity and space complexity, is widely used in various tasks. Neural networks are at the centre of the recent rise of AI and machine learning and have been used successfully in various fields such as image recognition and natural language processing. However, these models are capable of only local interpretation, such as feature importance, and do not have global interpretability.

Second, we introduce models that focus on interpretability. Linear regression is a classical model with high interpretability. Generalized linear models \cite{GLM} express the objective variable as a nonlinear transformation of a linear combination of features, which enables more flexible modelling. Generalized additive models \cite{GAM} are models that relax restrictions of the linear combination of generalized linear models, and the influence of each feature can be expressed as a nonlinear function. These models are highly interpretable, as the influence of each feature can be understood visually and intuitively. However, the prediction accuracy of the models is generally not very high, and they tend to be inaccurate for data having inherent higher-order interactions.

Finally, we introduce GA\textsuperscript{2}M, a model that achieves both high prediction accuracy and interpretability \cite{GA2M}. GA\textsuperscript{2}M is a generalized additive model with an additional interaction term. It is modelled as
\begin{equation}
y = \sum_{i \in Z^1} f_i(x_i)+\sum_{(ij)\in Z^2} f_{ij}(x_i,x_j)
\end{equation}
where $f_i(x_i)$ and $f_{ij}(x_i,x_j)$ are shape functions. $Z^1$ is an index set of all features, and we define $Z^2$ as $Z^2 := Z^{1} \times Z^{1}$. In this study, we used LightGBM as shape functions. Note that $i>j$ in the subscripts of the pairwise interaction terms. The pairwise interaction terms make it possible to model the interaction between the two features, and GA\textsuperscript{2}M is highly interpretable because the influence of each feature on the objective variable and the influence of the interaction between the two features can be visualized in a graph. In addition, the prediction accuracy is also high because nonlinear functions are used for each shape function and pairwise interactions are considered, rather than the effects of the features alone. In this study, we developed a model based on the idea of GA\textsuperscript{2}M that combines higher levels of prediction accuracy and interpretability. Another approach to an interpretation is to interpret the prediction results. For example, we can use LIME \cite{LIME} or SHAP \cite{SHAP} to interpret prediction results of a model like GBDT. However, these approach only provides local interpretations of the prediction results and cannot provide a global interpretation of the whole model. It should be noted that the purpose of this study is to interpret the whole model, which is fundamentally different from interpreting the prediction results by using LIME or SHAP.

\section{Proposed Method}
\subsection{Problems with GA2M}
\begin{figure}[b]
\centerline{\includegraphics[width=7cm]{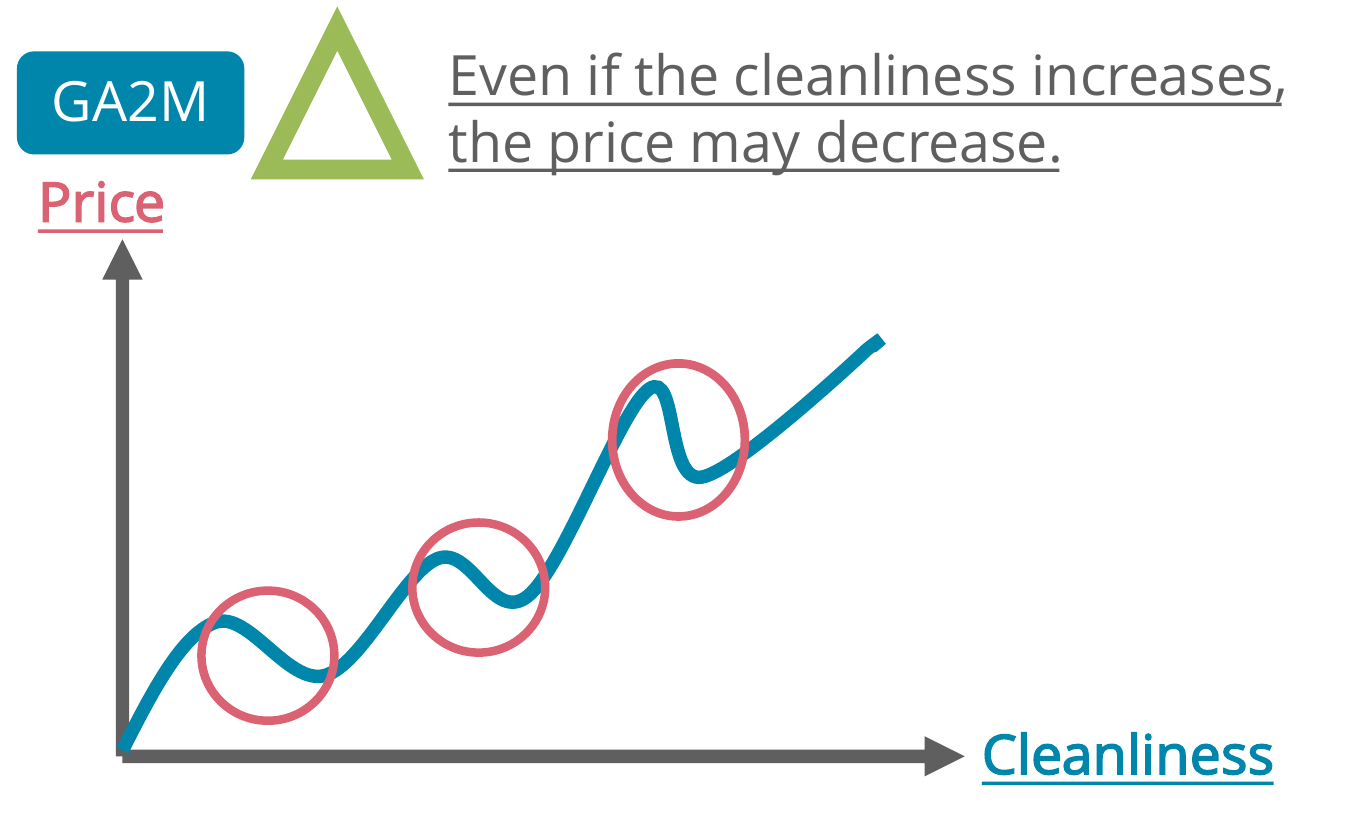}}
\caption{Example of the need to introduce monotonicity}
\label{fig1}
\end{figure}

GA\textsuperscript{2}M suffers from two major problems. The first one is the interpretability of the shape function. GA\textsuperscript{2}M can use any function as the shape function, and analysts can observe each shape function to infer how the individual feature affects the objective variable. However, the shape function may show illogical results due to noise in the data or lack of data. For example, consider the relationship between the housing prices and the number of rooms of it. In general, as the number of rooms increases, the real estate price rises. However, if we simply apply GA\textsuperscript{2}M to the data, due to the aforementioned causes, the result may be that the housing prices decreases even though the number of rooms increases, as shown in Fig. \ref{fig1}. This is not logically correct but also does not provide useful suggestions to analysts. Therefore, we apply monotonic constraints to some shape functions of a model to prevent it from producing illogical results. Returning to the previous example, we force the model to learn the relationship that housing prices increase as the number of rooms increases. We explain the details of this method in Section \ref{sec:propose}. The second problem is that GA\textsuperscript{2}M can not capture higher-order interactions. The shape functions in GA\textsuperscript{2}M include univariate terms and pairwise interaction terms but cannot consider more higher-order interactions, resulting in lower accuracy than models that can account for higher-order interactions. For example, higher-order interactions such as $x_1x_2x_3x_4$ difficult to approximate with a GA\textsuperscript{2}M. Therefore, we improved GA\textsuperscript{2}M by considering higher-order interactions, which do not have interpretability. Further, we devised a learning method that does not interpretability of the whole model.

\subsection{Constrained Generalized Additive 2 Model plus}\label{sec:propose}
We propose CGA\textsuperscript{2}M+, which solves the two abovementioned problems of GA\textsuperscript{2}M. CGA\textsuperscript{2}M+ is modelled as follows:

\begin{equation}
\label{eq:cga2m}
\begin{split}
    y =& \sum_{i \in Z_c}f_i(x_i)+ \sum_{i \in Z_u}f_i(x_i)+\sum_{(ij) \in Z_{cc}} f_{ij}(x_i,x_j)  \\& +\sum_{(ij) \in Z_{cu}}f_{ij}(x_i,x_j)+\sum_{(ij) \in Z_{uu}}f_{ij}(x_i,x_j) \\
    &+f_{\mathrm{high}}(x_1,x_2,\dots,x_K)
\end{split}
\end{equation}
The symbols in Eq (\ref{eq:cga2m}) are as follows.
\begin{itemize}
    \item $Z_c$ : an index set of features which are enforced monotonicity constraints on.
    \item $Z_u$ : an index set of features which are not enforced monotonicity constraints on.
    \item $Z_{cc}$ : $Z_{cc} = Z_c \times Z_c$
    \item $Z_{cu}$ : $Z_{cu} = Z_c \times Z_u$
    \item $Z_{uu}$ : $Z_{uu} = Z_u \times Z_u$
\end{itemize}

During training CGA\textsuperscript{2}M+, feature pairs are selected from a set $Z^2(=Z_{cc}\cup Z_{cu}\cup Z_{uu})$. 

CGA\textsuperscript{2}M+ solves the problem of interpretability of shape functions in GA\textsuperscript{2}M. We train the model by enforcing monotonicity in the corresponding shape functions based on human knowledge, such as the fact that housing prices increase as the number of rooms increases. We chose LightGBM as the shape function because of its fast training speed and high accuracy. we can impose monotonicity constraints on any shape functions if necessary. The monotonicity constraint is implemented by imposing a constraint on decision tree branching. Specifically, we penalize branches that violate the monotonicity constraint. In addition, the branching is performed in compliance with the monotonicity constraint by transferring the constraint of the parent node to the child node. For details, refer to [12].

CGA\textsuperscript{2}M+ also solves the problem of GA\textsuperscript{2}M’s inability to capture higher-order interactions. To capture higher-order interactions without losing interpretability, we added the function $f_{\mathrm{high}} (x_1,x_2,...x_K)$ (called the higher-order term) to GA\textsuperscript{2}M. This function is not interpretable but can capture higher-order interactions, and the proposed learning method does not reduce interpretability of the whole model.

\begin{figure}[t]
\centerline{\includegraphics[width=\linewidth]{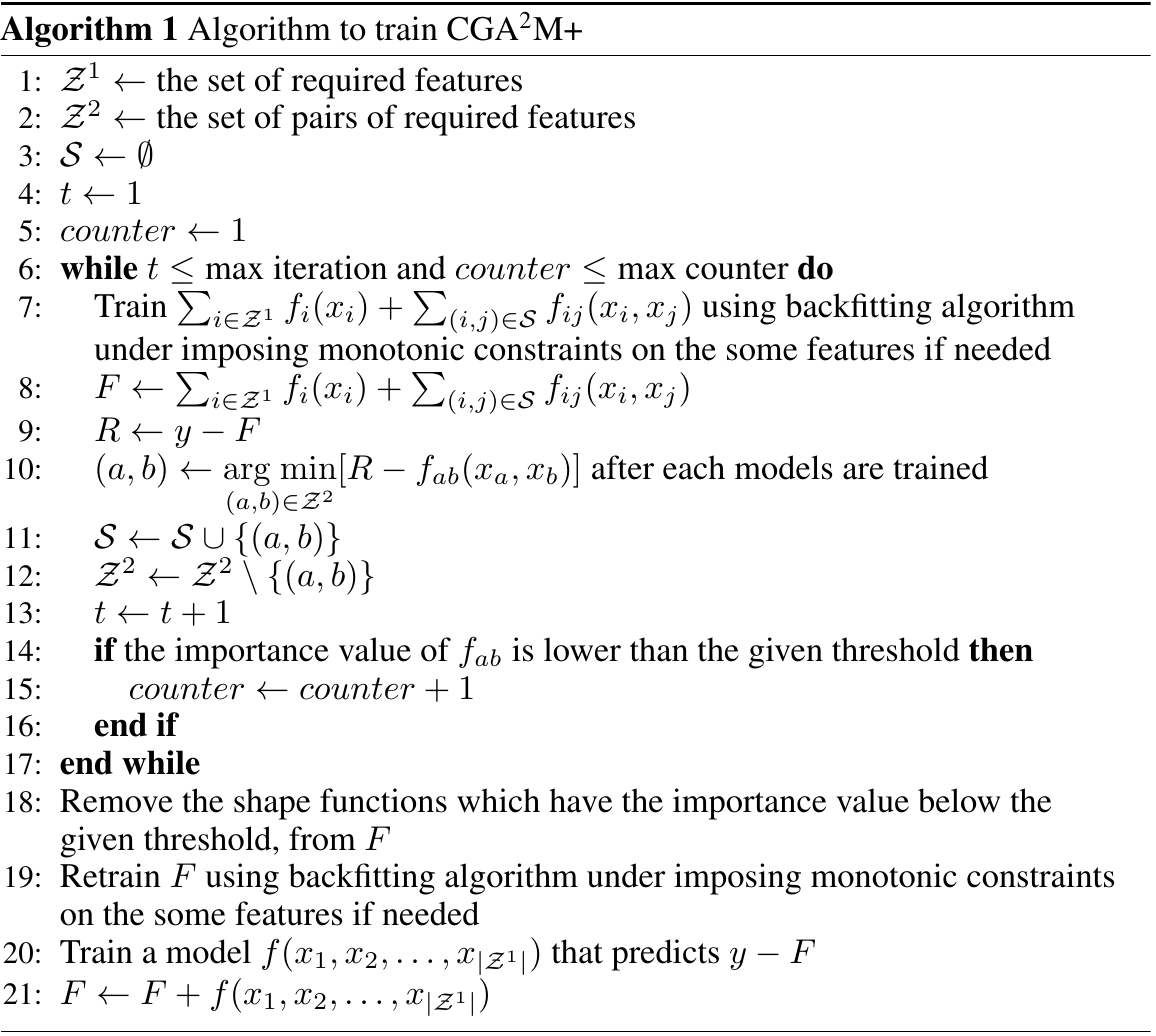}}
\end{figure}

The training of the proposed CGA\textsuperscript{2}M+ is performed as in Algorithm 1. There are two points to note in the algorithm. First, the pair of features in $Z^2$ are not necessarily used for training. $Z^2$ is just a set of feature pairs that can be selected, and we select the pairs from this set sequentially. Therefore, unimportant feature pairs are deleted from the model in line 14. The next point is that $f_{\mathrm{high}}$ does not affect the learning of $f_i$ and $f_{ij}$. As can be seen in line 20, $f_i$ and $f_{ij}$ have already been learned when learning $f_{\mathrm{high}}$, and therefore they are unaffected. In other words, the interpretability of $f_i$  and $f_{ij}$ is ensured by this learning mechanism.
Next, we define importance of shape functions and the higher-order term. In the training of CGA\textsuperscript{2}M+, lines 14 and 18 of Algorithm 1 determine whether the shape function is effective, and the method is explained below. Assume that $F$ is a trained CGA\textsuperscript{2}M+:

\begin{equation}
    F=\sum_{i} f_{i}\left(x_{i}\right)+\sum_{i<j} f_{i j}\left(x_{i}, x_{j}\right)+f_{\mathrm{high}}\left(x_{1}, x_{2}, \ldots x_{K}\right)
\end{equation}

Let $x_{in}$ be the value of the $i$th feature $x_i$ in the $n$th $(n \in \{1,2,...,N\})$ training sample. Also, let the value of the objective variable in the $n$th training data be $y_n$. The importance of $f_i$, $f_{ij}$, and $f_{\mathrm{high}}$ is defined as

\begin{align}
    &\mathrm{importance ~ of}~ f_i=\frac{\mathrm{effect}_i}{\mathrm{effect_{all}}}\\
    &\mathrm{importance ~ of}~ f_{ij}=\frac{\mathrm{effect}_{ij}}{\mathrm{effect_{all}}}\\
    &\mathrm{importance ~ of}~ f_{\mathrm{high}}=\frac{\mathrm{effect_{high}}}{\mathrm{effect_{all}}}
\end{align}

where the respective definitions of $\mathrm{effect}_i$,$\mathrm{effect}_{ij}$, $\mathrm{effect_{high}}$ and $\mathrm{effect_{all}}$ are

\begin{align}
    & \mathrm {effect}_{i}=\frac{\sum_{n=1}^{N}\left|f_{i}\left(x_{in}\right)-\bar{f}_{i}\right|}{\sum_{n=1}^{N}\left|y_{n}-\bar{y}\right|}\\
    & \mathrm {effect}_{ij}=\frac{\sum_{n=1}^{N}\left|f_{ij}\left(x_{in}, x_{jn}\right)-\bar{f}_{ij}\right|}{\sum_{n=1}^{N}\left|y_{n}-\bar{y}\right|,}\\
    & \mathrm{effect}_{\mathrm{high}}=\frac{\sum_{n=1}^{N}\left|f_{\mathrm{high}}\left(x_{1n}, x_{2n}, \ldots x_{Kn}\right)-\bar{f}_\text{high}\right|}{\sum_{n=1}^{N}\left|y_{n}-\bar{y}\right|,}\\
    &\mathrm{effect}_{\mathrm{all}}=\sum_{i} \mathrm{ effect}_{i}+\sum_{ij} \mathrm{effect}_{ij}+\mathrm{effect}_{\mathrm{high}}
\end{align}

In (7) to (10), we use the following symbols.

\begin{align}
    &\bar{f}_i=\frac{1}{N}\sum_{n=1}^{N}f_i(x_{in})\\
    &\bar{f}_{ij}=\frac{1}{N}\sum_{n=1}^{N}f_{ij}(x_{in},x_{jn})\\
    &\bar{f}_{\mathrm{high}}=\frac{1}{N}\sum_{n=1}^{N}f_{\mathrm{high}}(x_{n1},x_{n2},\dots,x_{nK})\\
    &\bar{y}=\frac{1}{N}\sum_{n=1}^{N}y_{n}
\end{align}

In line 18 of Algorithm 1, shape functions with importance below a given threshold are deleted. As described in Section \ref{sec:propose}, the training method in Algorithm 1 controls the importance of the higher-order term, which is not interpretable. Therefore, the accuracy of the model can be improved without reducing interpretability of the whole model. In this study, the threshold is set to 0.01.

\section{Numerical Experiment}
We next introduce the datasets used in the experiments, evaluate the proposed monotonicity and higher-order term, compare the proposed method with existing methods, and finally outline some application examples.
\subsection{Datasets}
We conducted experiments on the two datasets described in Table \ref{tab:dataset}.

\begin{table}[b]
    \caption{Dataset}
    \label{tab:dataset}
    \centering
    \begin{tabular}{|c|c|c|c|}
    \hline
\textbf{Data}&\textbf{Sample Size}&\textbf{Features}&\textbf{Monotonicity}\\
    \hline
    California housing prices&20,640&9&3\\
    \hline
    Oricon cafe satisfaction&6,537&37&31\\
    \hline
    \end{tabular}
\end{table}

``California housing prices'' is a dataset obtained from Kaggle \cite{california}, and it contains data about housing prices in various regions of California. ``Oricon cafe satisfaction'' is a dataset from a customer satisfaction survey about cafes conducted by Oricon Inc. We used the Oricon customer satisfaction survey data provided in the 2020 Data Analysis Competition sponsored by the Joint Association Study Group of Management Science \cite{compe}. In the California housing prices dataset, the task is to predict the housing prices based on features such as annual income, age and so on. In the Oricon cafe satisfaction dataset, the task is to predict how satisfied a customer is with a particular cafe based on cafe features and customer features. In Table \ref{tab:dataset}, \textbf{Features} is the number of features and \textbf{Monotonicity} is the number of features for which monotonicity is introduced when CGA\textsuperscript{2}M+ is applied. The ratio of training, validation, and test data was set to 60\%, 20\%, and 20\%, for each dataset.

\begin{table}[b]
\centering
\caption{Evaluation of Monotonicity and Higher-Order Term}
\label{tab:eval}
\begin{tabular}{|l|c|c|c|c|}
\hline
{} & \multicolumn{2}{|c|}{\bf California} & \multicolumn{2}{|c|}{\bf Oricon}\\
\hline
\bf Model & \bf \textit{training} & \bf \textit{test} & \bf \textit{training} & \bf \textit{test}\\
\hline
\begin{tabular}{l}GA\textsuperscript{2}M \end{tabular}& 31,263.74 & 50,488.58 & 0.953 & 1.167\\ \hline
\begin{tabular}{l}GA\textsuperscript{2}M\\+higher \end{tabular}& 27,587.23 & 49,923.73 & 0.927 & 1.162 \\ \hline
\begin{tabular}{l}GA\textsuperscript{2}M\\+monotonicity\end{tabular} & 39,436.42 & 48,543.97 & 1.016 & 1.141 \\ \hline
\begin{tabular}{l}GA\textsuperscript{2}M\\+ monotonicity\\+ higher\end{tabular} & 33,756.81 & \bf 47628.92 & 1.006 &\bf 1.136 \\ \hline
\end{tabular}
\end{table}

\subsection{Evaluating Monotonicity and Higher-Order Term}
Table \ref{tab:eval} shows the evaluation of the effect of introducing the monotonicity and the higher-order term proposed in this paper. We compare our proposed method, CGA\textsuperscript{2}M+, with other existing methods using the root mean squared error.
In both datasets, CGA\textsuperscript{2}M+ which introduced both monotonicity and a higher-order term shows the best results. We can also see that the introduction of the monotonicity reduced the test error, and as shown in Section \ref{sec:application}, improved interpretability. This suggests that the use of domain knowledge improves generalization performance. In contrast, although the effect of introducing the higher-order term is not as apparent with the Oricon data, some improvement in accuracy can be confirmed in the California housing data. The effectiveness of the higher-order term may vary depending on whether higher-order interactions are inherent in the data. The validity of interpreting CGA\textsuperscript{2}M+ with a non-interpretable higher-order term is discussed in Section \ref{sec:valid}. 

\begin{table}[b]
\centering
\caption{Comparison with Existing Method}
\label{tab:exist}
\begin{tabular}{|c|c|c|c|c|}
\hline
{} & \multicolumn{2}{|c|}{\bf California} & \multicolumn{2}{|c|}{\bf Oricon}\\
\hline
\bf Model & \bf \textit{training} & \bf \textit{test} & \bf \textit{training} & \bf \textit{test}\\
\hline
Linear regression& 69,706.48 & 69,037.95 & 1.207 & 1.210\\ \hline
Random Forest& 18,313.34 & 49,829.04 & 0.419 & 1.160 \\ \hline
LightGBM& 41,893.29 & 48,880.43 & 1.090 & 1.148 \\ \hline
Multi-Layer Perceptron& 68,616.29 & 68,454.99 & 1.171 & 1.290 \\ \hline
CGA\textsuperscript{2}M+& 33,756.81 & \bf 47628.92 & 1.006 & \bf 1.136 \\ \hline
\end{tabular}
\end{table}

\subsection{Comparison with Existing Methods}
The results of comparing the proposed CGA\textsuperscript{2}M+ with existing methods are shown in Table \ref{tab:exist}. The models are compared by the root mean square error.

To implement the existing methods, we used the scikit-learn library [14] for linear regression, random forest, and multi-layer perceptron, and the LightGBM library [12] for LightGBM. In addition, the default settings were used for the hyperparameters of each method. From Table \ref{tab:exist}, we can see that CGA\textsuperscript{2}M+ gave the best results for both datasets. Linear regression is not highly accurate due to its lack of high expressive power, but it does not overfit the model, and there is little difference between the performance of training and test data. Random forest shows a considerable difference in accuracy between the training and test data, indicating that the model is overfitted. In addition, multi-layer perceptrons often require tuning of hyperparameters, such as layer depth and number of units, but in this case, the default values were used, resulting in unlearning.

\begin{figure*}[tb]
  \begin{minipage}[b]{0.5\linewidth}
    \centering
    \includegraphics[width=\linewidth]{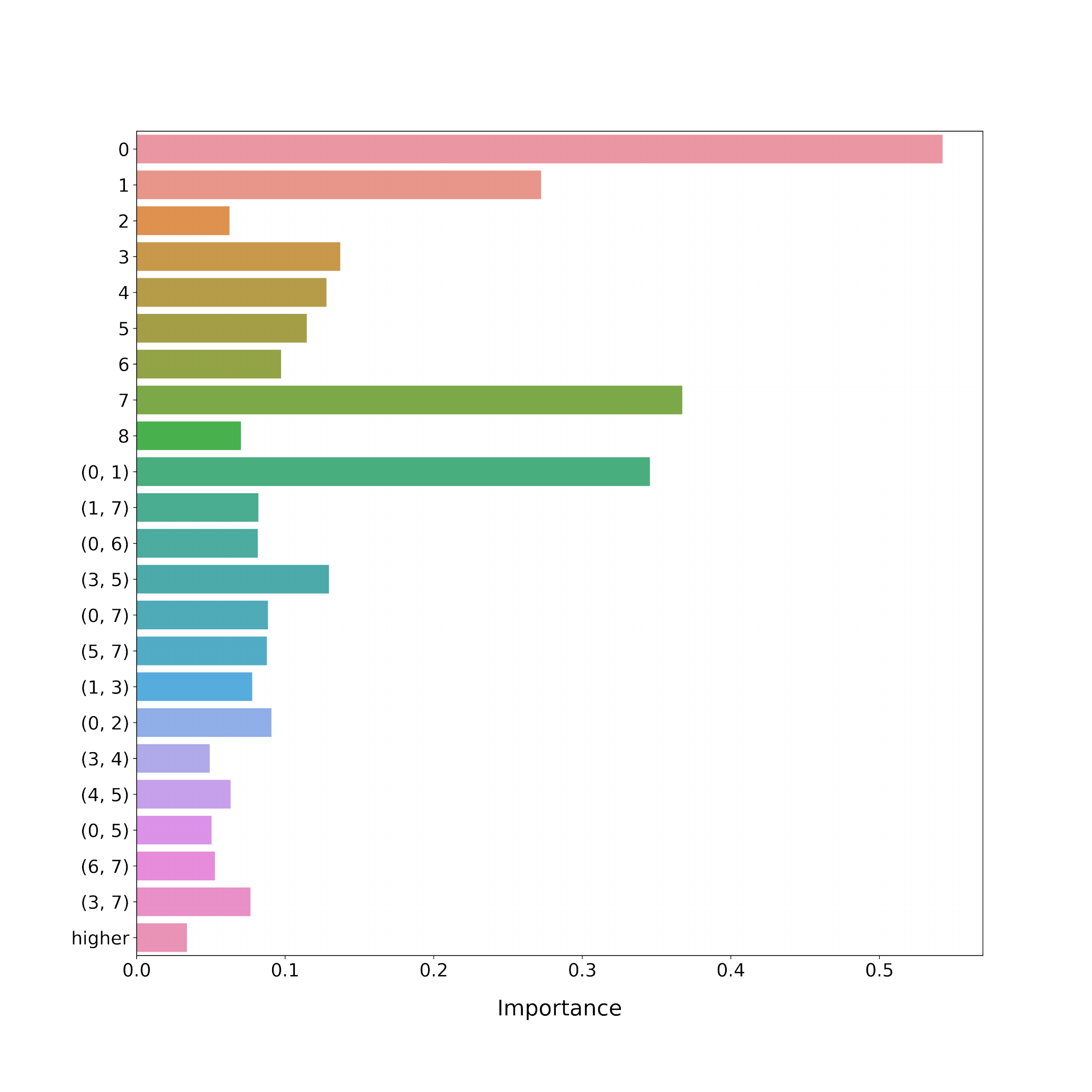}
    \subcaption{California}
  \end{minipage}
  \begin{minipage}[b]{0.5\linewidth}
    \centering
    \includegraphics[width=\linewidth]{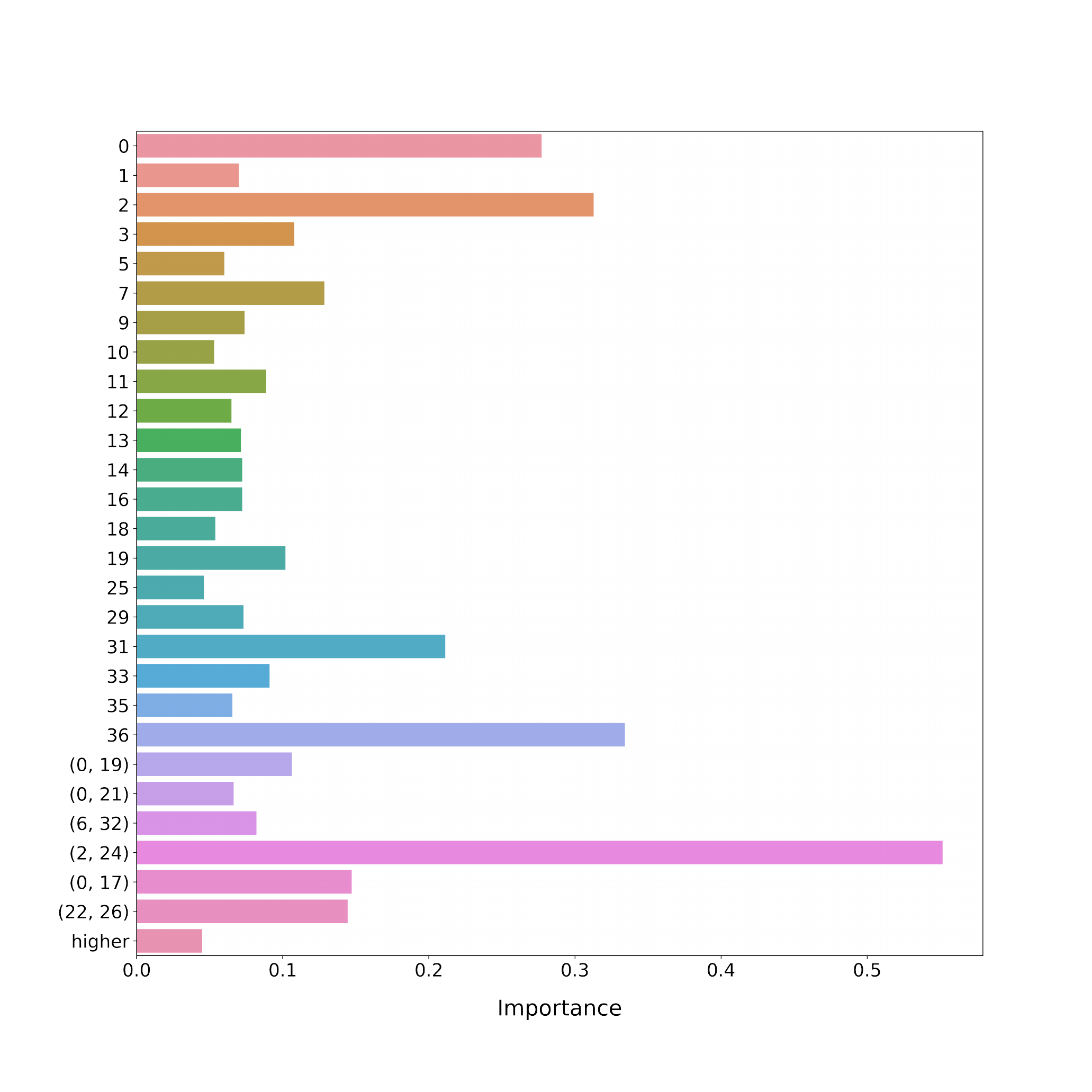}
    \subcaption{Oricon}
  \end{minipage}
  \caption{Feature importance for each dataset}
  \label{fig:importance}
\end{figure*}

\subsection{Validity of Model Interpretation} \label{sec:valid}
Fig. \ref{fig:importance} shows a bar graph showing the importance of the univariate terms, pairwise interaction terms, and the higher-order term introduced in Section \ref{sec:propose}. The vertical axis is the number assigned to each term. Some features that have low importance are omitted for clarity. The importance of the higher-order term (bottom of each plot) is not that great. This is because most of the data can be explained by univariate and pairwise interaction terms, and higher-order interactions are not inherent in the data. In such a case, the validity of interpreting the whole model is relatively high, and both prediction accuracy and interpretability can be achieved. Conversely, when the importance of the higher-order term is large, the parts that can be explained by the univariate and pairwise interaction terms are relatively small, and the interpretation validity of the whole model becomes low. In other words, the importance of the higher-order term can be used as a measure for the interpretation validity of the univariate and pairwise interaction terms.

\subsection{Applications}\label{sec:application}
In this section, we introduce some examples of applications that take advantage of the interpretability of CGA\textsuperscript{2}M+. Note that the figures in this section can be prepared for all the univariate and pairwise interaction terms used in the trained model, but only some of them are shown for saving the space. Here, although explanations are given only for the univariate terms, the pairwise interaction terms can be interpreting in the same way.

\begin{figure}[tb]
\centerline{\includegraphics[width=\linewidth]{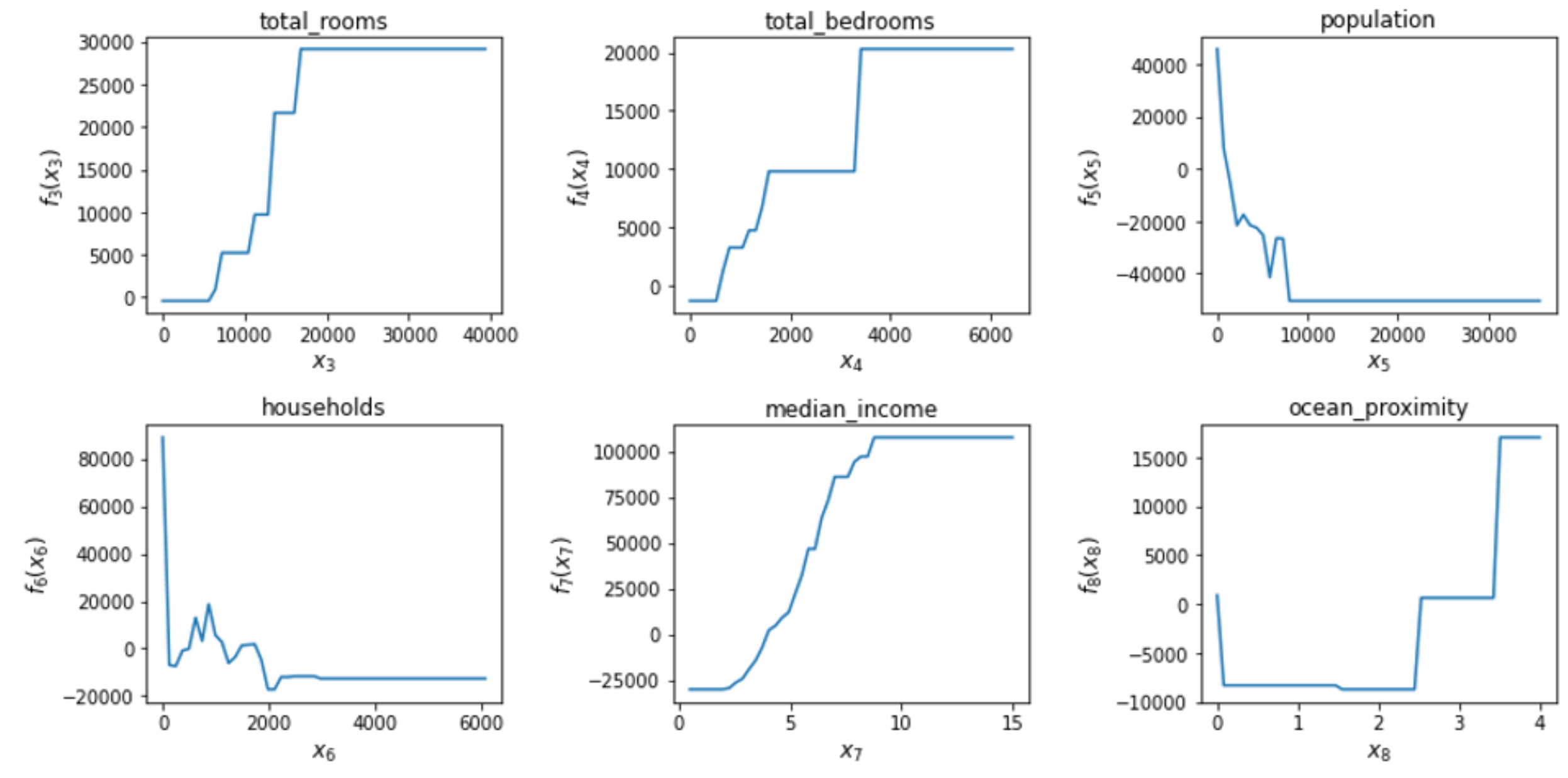}}
\caption{Plots of $f_i$ for California housing prices}
\label{fig:california}
\end{figure}

\paragraph{California Housing Prices}
Fig. \ref{fig:california} shows the relationship between each feature and the housing prices. We focus on the median income plot at the bottom middle. This is the median annual income in a certain region, and intuitively, the housing prices in that region is expected to increase monotonically with this value. Therefore, in this experiment, we imposed a monotonic increase constraint on this feature. The effect of the monotonic constraint shows that the housing prices monotonically increases. It can also be seen that the relationship between median income and housing prices is close to linear in the range of median income $<$ 10, and that median income does not affect housing prices when median income $\geq$ 10. Also, we imposed monotonic constraints on other features, such as average age and average number of rooms. These features can be interpreted in the same way as those of median income. We did not impose monotonicity on the other features, but if analysts have corresponding domain knowledge, they can consider introducing monotonicity to other features.

\begin{figure}[h]
\centerline{\includegraphics[width=\linewidth]{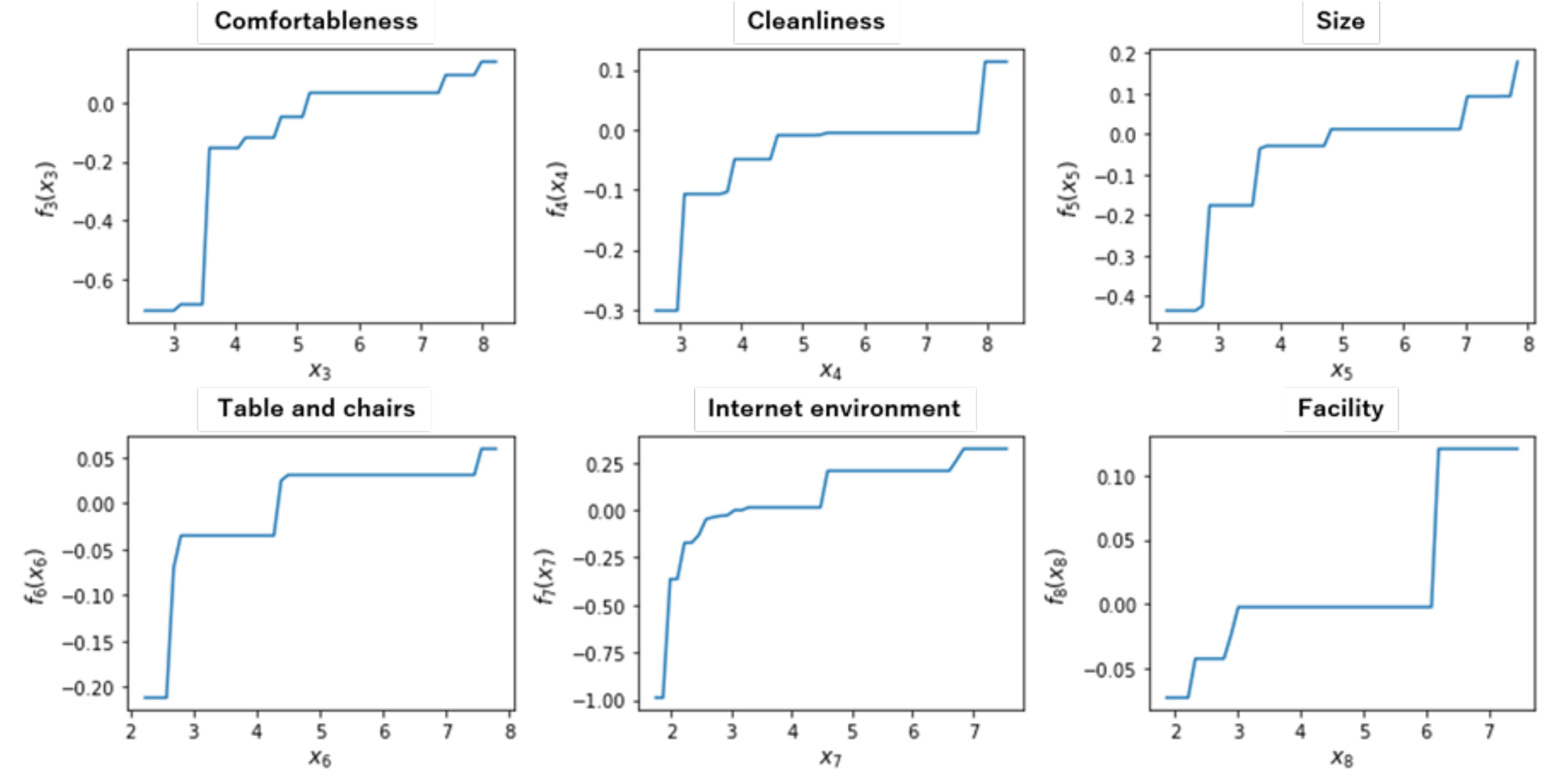}}
\caption{Plots of $f_i$ for Oricon cafe satisfactions}
\label{fig:oricon}
\end{figure}

\paragraph{Oricon Cafe Satisfactions}
In this paragraph, we discuss how the cleanliness score in a cafe affects the satisfaction score of the cafe. 
On the top center of Fig. \ref{fig:oricon}, when the cleanliness score for the cafe is 3.5, the cafe can increase the satisfaction score by 0.05, by improving the cleanliness score by 1. Also, the significantly lower Internet environment score has a large negative impact on the satisfaction score. This is probably because the Internet environment is an infrastructure element, and to some extent, it is taken for granted that it is installed. As a result, considering how the changes in each feature affect the objective variable leads to the interpretation of the whole model.

\section{Conclusion}
In this study, we proposed CGA\textsuperscript{2}M+, which has the high prediction performance and interpretability, by modifying the existing GA\textsuperscript{2}M model. The proposed model has two major improvements over GA\textsuperscript{2}M. The first is the use of domain knowledge, which improves the interpretability of the model and the generalization performance by imposing monotonicity constraints to some functions. The second is introduction of a higher-order term, which captures higher-order interactions that can not be explained by univariate and pairwise interaction terms, thereby improving the prediction accuracy.

In numerical experiments, we conducted a regression task using two datasets and confirmed that introducing the monotonicity and the higher-order term in CGA\textsuperscript{2}M+ is effective and that CGA\textsuperscript{2}M+ is superior to other models in terms of prediction accuracy. We also presented the two case studies, which demonstrate that CGA\textsuperscript{2}M+ has high interpretability.\\

\end{document}